\documentclass{article}

\PassOptionsToPackage{numbers, compress}{natbib}

\usepackage[preprint]{neurips_2023}

\usepackage{amsmath}
\usepackage{amssymb}

\usepackage[utf8]{inputenc} 
\usepackage[T1]{fontenc}    
\usepackage{hyperref}       
\usepackage{url}            
\usepackage{booktabs}       
\usepackage{amsfonts}       
\usepackage{nicefrac}       
\usepackage{microtype}      
\usepackage{xcolor}         

\usepackage{graphicx}
\newcommand{\tabincell}[2]{\begin{tabular}{@{}#1@{}}#2\end{tabular}}
\usepackage{multirow}
\usepackage{caption}
\usepackage{subcaption}
\usepackage{makecell}

\title{Auto-Spikformer: Spikformer Architecture Search}

\author{%
  \makecell{
Kaiwei Che$^{1,2}$, Zhaokun Zhou$^{1,2}$, Zhengyu Ma$^{2}$, Wei Fang$^{1,2}$, \\
Yanqi Chen$^{1,2}$, Shuaijie Shen$^{3}$, Li Yuan$^{1,2\, *}$, Yonghong Tian$^{1,2}$\thanks{Corresponding authors.}}\\
  $^{1}$ Peking University, China\\
  $^{2}$ Peng Cheng Laboratory, China\\
  $^{3}$ Southern University of Science and Technology, China\\
}

\begin{document}

\maketitle

\begin{abstract}

The integration of self-attention mechanisms into Spiking Neural Networks (SNNs) has garnered considerable interest in the realm of advanced deep learning, primarily due to their biological properties. Recent advancements in SNN architecture, such as Spikformer, have demonstrated promising outcomes by leveraging Spiking Self-Attention (SSA) and Spiking Patch Splitting (SPS) modules. However, we observe that Spikformer may exhibit excessive energy consumption, potentially attributable to redundant channels and blocks. To mitigate this issue, we propose Auto-Spikformer, a one-shot Transformer Architecture Search (TAS) method, which automates the quest for an optimized Spikformer architecture.
To facilitate the search process, we propose methods Evolutionary SNN neurons (ESNN), which optimizes the SNN parameters, and apply the previous method of weight entanglement supernet training, which optimizes the Vision Transformer (ViT) parameters. Moreover, we propose an accuracy and energy balanced fitness function $\mathcal{F}_{AEB}$ that jointly considers both energy consumption and accuracy, and aims to find a Pareto optimal combination that balances these two objectives.
Our experimental results demonstrate the effectiveness of Auto-Spikformer, which outperforms the state-of-the-art method including CNN or ViT models that are manually or automatically designed while significantly reducing energy consumption.
\end{abstract}

\section{Introduction}

Spiking neural networks (SNNs) are promising for the next generation of artificial intelligence due to their biological inspiration and appealing features such as sparse activation and temporal dynamics. The performance of SNNs has improved by utilizing advanced architectures from ANNs such as ResNet-like SNNs \cite{hu2018residual,fang2021deep, zheng2021going, hu2021advancing} or Spiking Recurrent Neural Networks \cite{lotfi2020long}. 
Transformer, originally developed for natural language processing\cite{vaswani2017attention}, has been successful in a variety of computer vision applications, including image classification\cite{dosovitskiy2020image,yuan2021tokens}, object detection \cite{carion2020end,zhu2020deformable,liu2021swin}, semantic segmentation \cite{wang2021pyramid,yuan2021volo}. The self-attention mechanism, a crucial component of the Transformer model, selectively attends to relevant information and is analogous to an important feature of the human biological system \cite{whittington2022relating,caucheteux2022brains}. The integration of self-attention into SNN for advanced deep learning has gained attention due to the biological properties of both mechanisms. Spikformer \cite{zhou2022spikformer}, a recent SNN architecture, has demonstrated promising results on both static and neuromorphic datasets using its Spiking Self-Attention (SSA) and Spiking Patch Splitting (SPS) modules. 

Although SNNs are known for their low energy consumption compared to ANN, our observations revealed that the energy consumption of Spikformer can be significantly reduced as it contains potentially redundant channels and blocks. Through experimentation, with only fewer blocks and channels, we obtain even better accuracy. It motivates us to search for improved Spikformer architectures that balance energy consumption and accuracy. Nevertheless, designing and training such hybrid models remains a challenging task \cite{dosovitskiy2020image, touvron2021training}. 
Transformer Architecture Search (TAS) \cite{chen2021glit, chen2021autoformer, su2022vitas} gains attention as an automated way to search for multiple configurations of Vision Transformer (ViT) architectures. 
The one-shot NAS scheme \cite{chen2021autoformer, dong2019one} is leveraged in TAS and obtains reliable performance estimations on various ViT architectures. 
The utilization of the one-shot NAS scheme, as proposed by \cite{chen2021autoformer, dong2019one}, is employed in the TAS and has proven to yield dependable performance estimations for diverse ViT architectures. Nevertheless, our experimental findings suggest that directly applying TAS may not be the most optimal solution for SNN. The original TAS method do not care about the SNN search space and the energy consumption, which is vital in the field of SNN.

To address this gap, we further analyze the two components of Spikformer: the Transformer architecture and SNN neurons. There are numerous prior methodologies proposed to optimize the Transformer architecture search, presenting various approaches that can be leveraged to achieve optimization. As to SNN neurons, the performance of an SNN neuron is determined by its internal parameters and interconnections. The Transformer component can also be regarded as the interconnections of SNN neurons in Spikformer. Several studies have focused on improving SNN performance by exploring the network's structure \cite{kim2022neural, che2022differentiable, na2022autosnn}. However, the internal parameters of individual neurons have also been identified as important. In this study, we propose a method to investigate the optimization of both the internal parameters of individual neurons and the interconnections between neurons. 
SNN neurons are mathematical models that approximate the behavior of biological neurons. Changes in the internal structure of biological neurons are derived from their adaptation to the environment, which is very similar to Darwin's theory of evolution \cite{jordan2021evolving, slowik2020evolutionary}, which suggests that living organisms adapt to their environment over time through natural selection. According to this theory, individuals with traits that are advantageous for survival in a particular environment are more likely to survive and reproduce, passing on their advantageous traits to their offspring.
In the context of SNN neurons, this concept of natural selection can be applied to the evolution of individual neurons. 
Specifically, the threshold $u_{th}$, decay $\tau$, and time-step $t$ parameters of a neuron can be regarded as its traits, while the input stimuli it receives can be considered as the environment in which it operates. Applying the concept of natural selection, the optimization of these parameter sets through simulated evolution may lead to improved network performance, resulting in increased accuracy and efficiency. Our study is the first to apply the evolutionary algorithm to search for the internal parameters of SNN neurons.


The energy consumption of Spikformer is influenced by several factors, such as the input image size, embedding dimension, number of blocks, firing rate, and time-step. Modifying the transformer architecture and selecting appropriate SNN parameters can adjust these factors. In evolutionary search, the fitness function is employed to evaluate candidate architectures. In order to achieve a balanced trade-off between energy consumption and accuracy, we incorporate energy as part of the fitness function. To the end, we propose a search space that considers both the original factors from the ViT and the additional factors from the SNN. Subsequently, we introduce a joint fitness function $\mathcal{F}_{AEB}$ that takes into account both energy consumption and accuracy in order to optimize this extended search space. This approach allows us to obtain a Pareto optimal combination of energy consumption and accuracy, striking an optimal balance between the two objectives.

In summary, our contributions are following:
\begin{itemize}
 \item To the best of our knowledge, this study is the first to use NAS for spiking-based ViT namely Auto-Spikformer. By using Evolutionary SNN neurons (ESNN) and weight entanglement \cite{chen2021autoformer} supernet training method, Auto-Spikformer enhances the efficiency and accuracy of spiking-based ViT architectures.
 
 \item Auto-Spikformer integrates an accuracy and energy balanced fitness function $\mathcal{F}_{AEB}$, to optimize the Spikformer search space by considering both energy consumption and accuracy simultaneously.

\item Auto-Spikformer has been successfully searched and evaluated on the CIFAR dataset, achieving state-of-the-art results when compared to other SNN models in both accuracy and energy consumption.
\end{itemize}

\section{Related work}

\subsection{Vision Transformer}
The Vision Transformer (ViT) facilitates the transformation from the NLP to the CV by partitioning visual information into patches and processing it accordingly.
Regarding the task of image classification, a Transformer encoder is comprised of a patch splitting module, multiple Transformer encoder blocks, and a linear prediction head. In each Transformer encoder block, there exists a self-attention layer along with a multi perception layer. Self-attention serves as the fundamental component contributing to the success of ViT. It enables the capture of global dependence and interest representation by weighing the feature values of image patches via the dot product of the query and key, followed by the application of the softmax function\cite{katharopoulos2020transformers,qin2022cosformer}. 
Researchers have made several improvements to the visual transformer, including the Transformer architecture\cite{xiao2021early,hassani2021escaping}, more advanced self-attention mechanisms\cite{song2021ufo,yang2021focal,rao2021dynamicvit,choromanski2020rethinking}, and pre-training techniques\cite{he2022masked}, among others.

\subsection{One Shot NAS}
Designing high performance network architectures for specific tasks often requires expert experience and trial-and-error experiments. Neural architecture search (NAS) \cite{elsken2019neural} aims to automate this manual process and has recently achieved highly competitive performance in tasks such as image classification \cite{zoph2016neural, zoph2018learning, liu2018progressive, real2019regularized, pham2018efficient}, object detection \cite{zoph2018learning, chen2019detnas, wang2020fcos, guo2020hit} and semantic segmentation \cite{liu2019auto, zhang2019customizable, nekrasov2019fast, lin2020graph}, etc. However, searching over a discrete set of candidate architectures often results in a massive number of potential combinations, leading to explosive computation cost.
The recently proposed differentiable architecture search (DARTS) method \cite{liu2018darts} and its variations \cite{xu2019pc, chen2019progressive, chu2020darts} address this problem using a continuous relaxation of the search space which enables learning a set of architecture coefficients by gradient descent, and has achieved competitive performances with the state-of-the-art using orders of magnitude fewer computation resources \cite{liu2018darts, liu2019auto, cheng2020hierarchical}. Recently, \cite{na2022autosnn} studied pooling operations for downsampling in SNNs and applied NAS to reduce the the overall number of spikes. \cite{kim2022neural} applied NAS to improve SNN initialization and explore backward connections. However, both works only searched for different SNN cells or combinations of them under fixed network backbone and their application is limited to image classification.

\subsection{Spiking Neural Networks}
Unlike traditional deep learning models that perform computations using floating-point values, SNNs leverage discrete spike sequences for information processing and transmission. Spiking neurons endow SNNs with temporal dynamics and biological properties, with common types including including the leaky integrate-and-fire (LIF) neuron \cite{wu2018spatio}, PLIF \cite{fang2021incorporating}, etc. There are two main approaches for obtaining deep SNNs: ANN-to-SNN conversion and direct training. In ANN-to-SNN conversion, a pre-trained ANN with high performance is transformed into an SNN by substituting the ReLU activation layers with spiking neurons\cite{cao2015spiking,hunsberger2015spiking,rueckauer2017conversion,bu2021optimal,meng2022training,wang2022signed}. However, this method requires large time-steps to approximate ReLU activation accurately, which leads to high latency \cite{han2020rmp}. In direct training, SNNs are trained by backpropagation through time (BPTT)\cite{werbos1990backpropagation}. A challenge for direct training is the non-differentiability of the event-triggered mechanism in spiking neurons. To address this challenge, surrogate gradients are employed for backpropagation \cite{lee2020enabling,neftci2019surrogate,xiao2021training} adopts implicit differentiation on the equilibrium state to train SNNs.

\section{Auto-Spikformer}

\subsection{LIF}
We adopt the iterative LIF neuron model \cite{wu2019direct} described by
\begin{equation}
    u^{t, n} = (1-\frac{1}{\tau}) u^{t-1,n}(1-y^{t-1,n}) + I^{t,n}
\label{eq1:LIF}
\end{equation}
where superscripts $n$ and $t$ denote layer index and time-step, respectively. decay $\tau$ is the membrane time constant,
$u$ is the membrane potential, $y$ denotes the spike output and $I$ denotes the synaptic input with 
$I^{t,n} = \sum_{j} w_{j} y^{t, n-1}_j$
where $w$ is the weight. The neuron will fire a spike $y^{t,n} = 1$ when $u^{t,n}$ exceeds a threshold $V_{th}$, otherwise $y^{t,n} = 0$. In this work, we set $\tau=2$ and $u_{th}=0.5$.

\begin{figure}[t!]
  \centering
  \includegraphics[width=1.\linewidth]{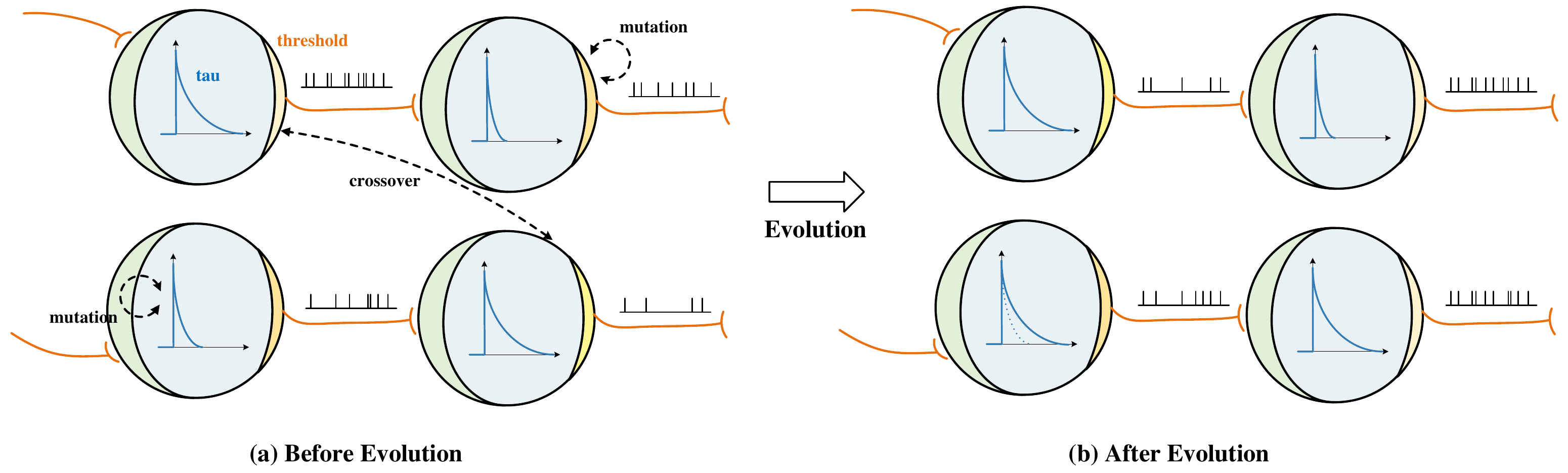}
  \caption{ESNN process. (a), (b) shows two candidate parameter sets before and after applying the mutation and crossover operators. The spike from the previous neuron is transmitted to the current neuron as the charging process. If the membrane potential is above the threshold (yellow area, where the darker color indicates a higher value), a spike is delivered, and if the membrane potential is below the threshold, it decays exponentially with a time constant $\tau$ (blue curve).}
  \label{fig-snn-space}
\end{figure}

\begin{table}[t!]
    \caption{Auto-Spikformer search space. $\mathcal{S}_{T_s}$ denotes transformer smaller search space, $\mathcal{S}_{T_l}$ denotes transformer larger search space, and $\mathcal{S}_{S}$ denotes SNN search space. Each element in the table represents the lower limit, upper limit and step size, such as depth (2,4,1) represents the search space is [2,3,4]}
    \label{tab-searchspace}
    \centering
    \fontsize{8.6pt}{9pt}\selectfont
    \begin{subtable}[t]{0.4\linewidth}
        \captionsetup{justification=centering}
        \begin{tabular}{cccc}
        \toprule
            & $\mathcal{S}_{T_s}$ & $\mathcal{S}_{T_l}$ \\
        \midrule
            embed dim & (336,384,12) & (336,480,48)\\
            MLP ratio & (3,4,0.2) & (3,5,0.2)\\
            head num & (6,12,6) & (6,12,6)\\
            depth & (2,4,1) & (2,6,1)\\
        \midrule
            
        \bottomrule
        \end{tabular}
        \caption{CIFAR search space}
    \end{subtable}
    \hspace{2.7cm}
    \centering
    \begin{subtable}[t]{0.26\linewidth}
        \captionsetup{justification=centering}
        \begin{tabular}{ccc}
        \toprule
            & $\mathcal{S}_{S}$  \\
        \midrule
            threshold $u_{th}$ & (0.6,2,0.2)\\
            decay $\tau$ & (1.25,10,0.25)\\
            time-step $t$ & (2,4,1)\\
        
        \bottomrule
        \end{tabular}
        \caption{SNN search space}
    \end{subtable}
\end{table}

\subsection{Evolutionary SNN Neurons (ESNN)}

The performance of SNN neurons is influenced by both their interconnections and internal parameters. While previous research has primarily focused on enhancing SNN performance through modifications to the network's structure, the importance of optimizing the internal parameters within individual neurons cannot be overlooked. 
Darwin's theory of evolution posits that organisms adapt to their surroundings through natural selection, favoring traits that enhance survival and reproduction. This concept can be applied to the context of SNN, where individual neurons can undergo an evolutionary process. In this context, the internal parameters of a neuron, such as the threshold ($u_{th}$), decay ($\tau$), and time-step ($t$), can be seen as analogous to traits, while the input stimuli received by the neuron can be likened to the environment in which it operates.

Previous work \cite{fontaine2014spike} suggests that the threshold can be viewed as an adaptation to membrane potentials at short timescales, influencing how signals received by a neuron are encoded into a spike. Decay $\tau$ has a similar effect to the threshold, but it only affects the decay of unfired neurons, influencing the firing of the next timestep. In contrast, the threshold affects the firing of all neurons at the current moment. 

As shown in Figure \ref{fig-snn-space}, the spike from the previous neuron is transmitted to the current neuron as the charging process. If the membrane potential is above the threshold, a spike is delivered, and if the membrane potential is below the threshold, it decays at the rate of $\tau$. The ESNN begins with a population of randomly generated parameter sets (candidates) like [$u_{th}=1.2$, $\tau=1.25$,$t=4$]. In each generation, the algorithm evaluates the fitness of candidates and selects the best ones as the parents for the next generation. The parents produce offspring by applying mutation and crossover operators with some probabilities. 
The mutation operator randomly modifies one parameter of a parameter set, while the crossover operator combines two parameters from different parents. As illustrated in Figure 2, for example, the decay $\tau$ of a candidate changes from 1.25 to 2.5 after mutation. the thresholds $u_{th}$ of two candidates are swapped after crossover, which affects the firing rate of each candidate.
The algorithm repeats this process for a fixed number of generations and returns the best architecture found.
Through a process of simulated evolution, the threshold, decay, and time-step parameters of individual neurons can be adjusted to improve the performance of the network as a whole. According to our experiments, this approach can lead to the network becoming better adapted to the input stimuli it receives, resulting in increased accuracy and efficiency.

Specifically, we design a SNN search space denoted $\mathcal{S}_{S}$ that includes three variable factors: the threshold $u_{th}$, decay $\tau$, and time-step $t$. The structured definition of this search space is outlined in Table \ref{tab-searchspace} (b), and its visual interpretation is depicted in Figure \ref{fig-snn-space}. To conduct this exploration, we adopt a supernet-based approach inspired by one-shot NAS methods, where all subnets share the same weights as the supernet, and only the SNN variable factors differ. Our process begins with the generation of an initial set of candidate parameter sets using a random search.  More details about the selection process are in the supplement.



\subsection{Accuracy and Energy Balanced Fitness Function ($\mathcal{F}_{AEB}$)}

In this study, we apply a large transformer search space that includes four variable factors to increase model capacities, similarly to the Autoformer \cite{chen2021autoformer}. We encode the search space into a supernet. However, the original fitness function only takes accuracy into account. We propose a new fitness function $\mathcal{F}_{AEB}$ that can balance the accuracy and the energy consumption. To estimate the energy use, we need to compute the synaptic operations (SOPs) first. For a specific layer $l$, its SOPs can be calculated as follows,

\begin{align}
        \mathrm{SOPs}(l)= fr \times t \times \mathrm{FLOPs}(l)  \label{eq:sop}
\end{align}
where $fr$ denotes the firing rate of the input spike train, $t$ represents time-step. Floating Point Operations (FLOPs) refer to the number of multiply-and-accumulate (MAC) operations. And SOPs contain spike-based accumulate (AC) operations only. We estimate the theoretical energy consumption of Auto-Spikformer according to \citep{kundu2021hire,hu2021advancing,horowitz20141,kundu2021spike,yin2021accurate,panda2020toward,yao2022attention}. We assume that the MAC and AC operations are implemented on the $45 \mathrm{nm}$ CMOS technology \cite{rathi2021diet}, where $E_{MAC}=4.6pJ$ and $E_{AC}=0.9pJ$. The theoretical energy consumption of Auto-Spikformer is calculated:

\begin{align}
    \mathcal{E}
     &= E_{MAC}  \times {\rm FL}^1_{{\rm SNN~Conv}} \nonumber \\  
     &+ E_{AC} \times \left(\sum_{n=2}^{N}{\rm SOP}^n_{{\rm SNN~Conv}} + \sum_{m=1}^{M}{\rm SOP}^m_{{\rm SNN~FC}} + \sum_{l=1}^L{\rm SOP}^l_{{\rm SSA}}\right) \label{eq:flop}
\end{align}
where $\mathcal{E}$ denotes the model energy, ${\rm FL}^1_{SNN~Conv}$ is the first layer to encode static RGB images into spike-form. Then the SOPs of $m$ SNN Conv layers, $n$ SNN Fully Connected Layer (FC) and $l$ SSA are added together and multiplied by $E_{AC}$. For ANNs, the theoretical energy consumption of block $b$ is calculated:
\begin{align}
    \mathrm{Power}(b)= 4.6{pJ} \times \mathrm{FLOPs}(b)  
    \label{eq:sop2}
\end{align}
For SNNs, $\mathrm{Power}(b)$  is:
\begin{align}
    \mathrm{Power}(b)= 0.9{pJ} \times \mathrm{SOPs}(b)  
    \label{eq:sop3}
\end{align}


The energy consumption of Spikformer is determined by multiple factors, including input image size, embedding dimension, number of blocks, firing rate $fr$, and time-step $t$. Most of these factors can be adjusted by changing the transformer architecture and selecting suitable spike neuron parameters. To facilitate comparison, we normalized these factors using a minmax scaler and assigned different weights to both metrics, the accuracy and energy balanced fitness function $\mathcal{F}_{AEB}$ is described as follows.

\begin{equation}
    \mathcal{F}_{AEB} = \alpha \times \mathcal{E} + (1-\alpha) \times \mathcal{A}
\end{equation}
where $\mathcal{A}$ denotes the top-1 accuracy. Both of them are scaled by a minmax scaler and the range is (0,1). $\alpha$ denotes the weight. In our case, we set $\alpha$ to 0.5.

\section{Experiments}

In this section, we provide a comprehensive overview of the implementation details and the settings used for the evolution search. Initially, we conduct an analysis to assess the effectiveness of the ESNN  by focusing on modifications within the SNN search space. Subsequently, we evaluate the efficacy of the $\mathcal{F}_{AEB}$ by comparing it with a random search approach and construct the Pareto frontier. Lastly, we present the performance evaluation of Auto-Spikformer on CIFAR dataset, providing comparisons with state-of-the-art models to highlight its effectiveness.

\subsection{Implementation Details}

Auto-Spikformer includes two stages: the supernet training stage and the evolutionary search stage. The implementation of all models utilized PyTorch 1.8 and the training process was performed on Nvidia Tesla V100 GPUs. Additional implementation details can be found in the supplementary materials.
During the supernet training stage, we followed a similar training approach as Spikformer, but with an extended epoch duration of 1000 to ensure improved convergence of the supernet. Subsequently, in the evolutionary search stage, we adopted a similar protocol as SPOS and Autoformer for the implementation of the evolutionary search. Our approach involved employing ESNN to explore thousands of candidate architectures and SNN parameter sets. We conduct our experiments on CIFAR dataset. \textbf{CIFAR} provides $50,000$ train and $10,000$ test images with $32\times 32$ resolution. The batch size is set to $128$.

\begin{figure}[t!]
  \centering
  \includegraphics[width=0.8\linewidth]{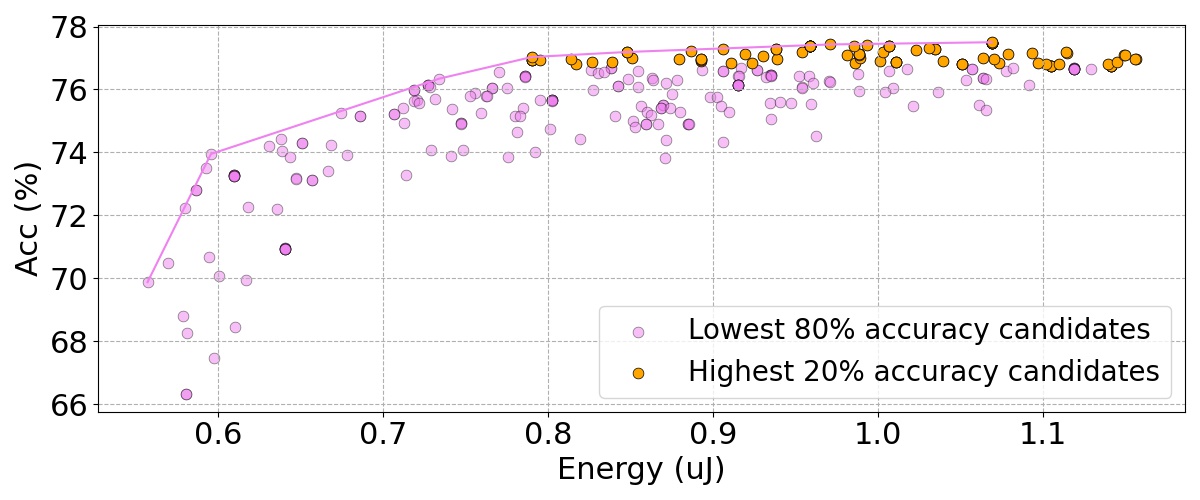}
  \caption{The energy and accuracy of all candidates in $\mathcal{S}_{S}$ in CIFAR100. We use $\mathcal{F}_{AEB}$ as the fitness function to select the top 300 candidates. The purple points represent the candidates with the lowest 80\% accuracy, while the orange points represent the candidates with the highest 20\% accuracy. The purple line represents the Pareto frontier, indicating the optimal trade-off between accuracy and energy consumption.}
  \label{fig-acc-eng}
\end{figure}

\begin{table}[t!]
    \caption{Subsets of the candidates in $\mathcal{S}_{S}$. It should be noted that the last row corresponds to Spikformer 4-384, which was obtained by running the open-source code of Spikformer in Auto-Spikformer mode. Therefore, its weight differs from the other models in the table. The bold font represents that the energy consumption and accuracy are superior to Spikformer 4-384.}
    \label{tab-snnspace_cand}
    \fontsize{7pt}{8pt}\selectfont
    \begin{subtable}[t]{0.4\linewidth}
    \centering
        \captionsetup{justification=centering}
        \begin{tabular}{cccc}
        \toprule
             \tabincell{c}{Candidates\\ (threshold $\times 4$,tau $\times 4$, time-step)} & Fr & \tabincell{c}{Energy \\($\mu J$)}  & \tabincell{c}{Acc \\ (\%)}\\
        \midrule
(1.6, 0.6, 0.8, 2.0, 10, 10, 10, 2, 2) & 0.20 & 0.52 & 72.12 \\
(1.8, 0.8, 1.4, 1.2, 10, 10, 10, 2, 2) & 0.20 & 0.52 & 72.18 \\
(1.6, 0.6, 2.0, 1.8, 5, 10, 10, 1.5, 4) & 0.24 & 0.63 & 75.58 \\
(1.0, 1.0, 1.4, 1.0, 10, 10, 2, 3, 4) & 0.25 & 0.66 & 76.44 \\
(1.8, 1.6, 0.6, 0.8, 5, 10, 2, 3, 4) & 0.26 & 0.68 & 76.77 \\
(0.8, 1.2, 1.6, 2.0, 5, 10, 2, 1.5, 4) & 0.27 & 0.72 & 77.20 \\
(1.0, 2.0, 1.6, 2.0, 5, 2, 2, 3, 4) & 0.30 & \textbf{0.79} & \textbf{77.87} \\
(1.0, 2.0, 1.4, 1.6, 2, 2, 1.25, 5, 4) & 0.37 & 0.99 & 77.95 \\

        \midrule
     (1.0, 1.0, 1.0, 1.0, 2, 2, 2, 2, 4) & 0.35 & 0.95 & 77.86 \\
     
        \bottomrule
        \end{tabular}
        \caption{Candidates on the Pareto frontier.}
    \end{subtable}
    \hspace{1.5cm}
    \begin{subtable}[t]{0.4\linewidth}
        \captionsetup{justification=centering}
        \begin{tabular}{cccc}
        \toprule
             \tabincell{c}{Candidates\\ (threshold $\times 4$,tau $\times 4$, time-step)} & Fr & \tabincell{c}{Energy \\($\mu J$)}  & \tabincell{c}{Acc \\ (\%)}\\
        \midrule

(1.0, 2.0, 1.6, 2.0, 5, 2, 2, 3, 4) & 0.30 & \textbf{0.79} & \textbf{77.87} \\
(1.2, 1.8, 1.8, 1.6, 2, 10, 1.5, 5, 4) & 0.33 & \textbf{0.87} & \textbf{77.90} \\
(1.6, 1.2, 1.8, 2.0, 1.5, 5, 10, 2, 4) & 0.33 & \textbf{0.88} & \textbf{77.86} \\
(1.6, 0.8, 1.2, 1.8, 2, 10, 1.25, 3, 4) & 0.34 & 0.91 & 77.74 \\ 
(0.6, 1.4, 1.2, 0.6, 2, 3, 1.25, 5, 4) & 0.35 & 0.94 & 77.78 \\ 
(0.8, 0.8, 1.8, 1.8, 2, 2, 1.5, 2, 3) & 0.36 & \textbf{0.95} & \textbf{77.92} \\ 
(1.0, 1.4, 1.8, 0.6, 2, 2, 1.5, 3, 4) & 0.36 & 0.96 & 77.77 \\ 
(1.0, 2.0, 1.4, 1.6, 2, 2, 1.25, 5, 4) & 0.37 & 0.99 & 77.95 \\ 
        \midrule
     (1.0, 1.0, 1.0, 1.0, 2, 2, 2, 2, 4) & 0.35 & 0.95 & 77.86 \\
     
        \bottomrule
        \end{tabular}
        \caption{Candidates with the top 20\% accuracy.}
    \end{subtable}

\end{table}

\begin{figure}[t!]
  \centering
  \begin{subfigure}[b]{0.495\textwidth}
    \includegraphics[width=\textwidth]{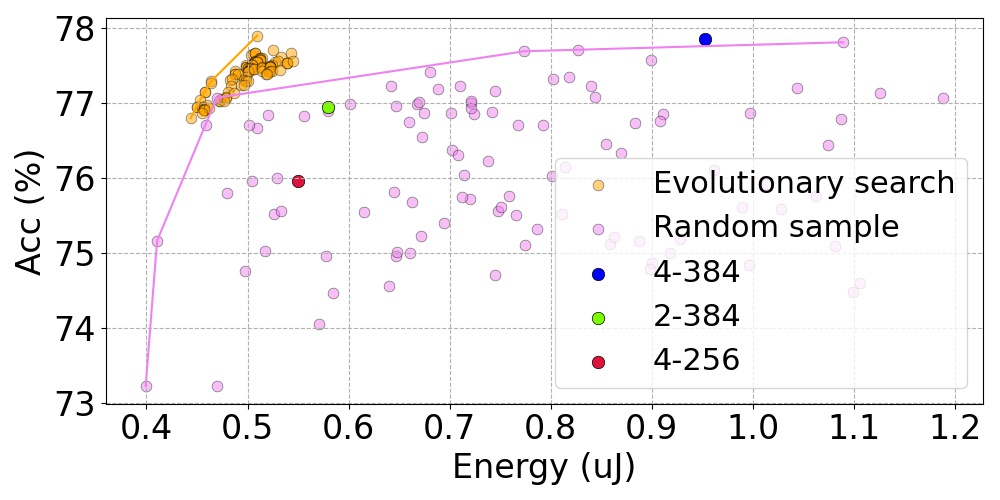}
    \caption{Candidates in $\mathcal{S}_{T_s}$.}
    \label{fig:esnn_snn_original_sahd}
  \end{subfigure}
  \hfill
  \begin{subfigure}[b]{0.495\textwidth}
    \includegraphics[width=\textwidth]{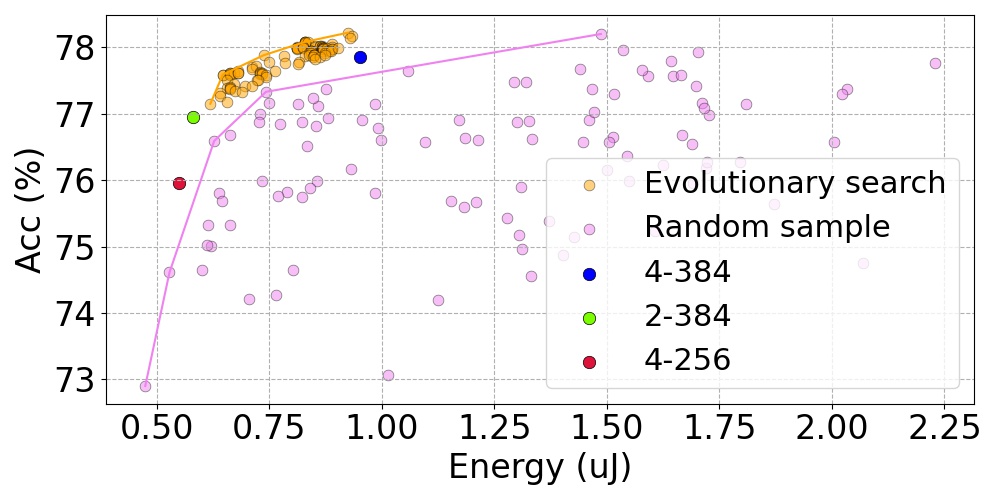}
    \caption{Candidates in $\mathcal{S}_{T_l}$.}
    \label{fig:esnn_snn_our_sahd}
  \end{subfigure}
  \caption{The energy and accuracy of all searched candidates in $\mathcal{S}_{T_s}$ and $\mathcal{S}_{T_l}$ in CIFAR100. We use $\mathcal{F}_{AEB}$ as the fitness function (purple points) and randomly (orange points) select the 100 candidates, respectively. The other color points represent the different architectures of Spikformer, which are derived from the original paper \cite{zhou2022spikformer}.}
  \label{fig-stspace-ae}
\end{figure}

\begin{table}[t!]
    \caption{Subsets of the candidates in $\mathcal{S}_{S}$. The bold font represents that the energy consumption and accuracy are superior to Spikformer 4-384.}
    \label{tab-stspace_cand}
\setlength{\tabcolsep}{2.7pt}
    \fontsize{6.5pt}{8pt}\selectfont
    \begin{subtable}[t]{0.4\linewidth}
    \caption{Candidates in $\mathcal{S}_{T_s}$. We denotes $^{1,2}$ as Auto-Spikformer $\mathcal{S}_{T_s1,2}$.}
        \captionsetup{justification=centering}
        \begin{tabular}{cccc}
        \toprule
             & \tabincell{c}{Candidates\\ (depth (d), MLP ratio $\times$ d, head num $\times$ d,\\ threshold $\times$ d, tau $\times$ d, time-step, embed dim)} & \tabincell{c}{Energy \\($\mu J$)}  & \tabincell{c}{Acc \\ (\%)}\\
        \midrule
\multicolumn{1}{c}{\multirow{5}{*}{\tabincell{c}{Pareto \\frontier}}} 
& (2, 3.2, 3.0, 12, 6, 1.2, 1.0, 5, 5, 4, 348) & 0.448 & 76.89 \\
& (2, 3.4, 3.2, 12, 6, 1.0, 1.0, 5, 5, 4, 348) & 0.453 & 77.04 \\
& (2, 3.8, 3.2, 12, 6, 0.6, 1.8, 5, 5, 4, 348) & 0.458 & 77.15 \\
& (2, 3.8, 3.8, 12, 6, 1.0, 1.8, 5, 5, 4, 348) & 0.464 & 77.29 \\
& (2, 3.6, 3.6, 6, 12, 1.8, 2.0, 5, 2, 4, 348) & \textbf{0.509} & \textbf{77.91} \\
        \midrule
\multicolumn{1}{c}{\multirow{5}{*}{\tabincell{c}{Top \\accuracy}}} 
& (2, 3.8, 3.6, 12, 12, 1.4, 2.0, 5, 2, 4, 348)$^1$ & 0.505 & 77.71 \\ 
& (2, 3.6, 3.6, 6, 12, 1.8, 2.0, 5, 2, 4, 348)$^2$ & \textbf{0.509} & \textbf{77.91} \\ 
& (2, 3.6, 3.6, 6, 12, 0.6, 2.0, 5, 2, 4, 348) & \textbf{0.510} & \textbf{77.91} \\ 
& (2, 3.4, 3.8, 6, 6, 1.0, 1.8, 5, 2, 4, 360) & \textbf{0.535} & \textbf{77.89} \\ 
& (2, 3.6, 3.6, 6, 6, 0.6, 2.0, 5, 2, 4, 360) & \textbf{0.536} & \textbf{77.90} \\ 
        \midrule
        
\tabincell{c}{Spikformer \\4-384}
& \tabincell{c}{(4, 4, 4, 4, 4, 12, 12, 12, 12,\\1.0, 1.0, 1.0, 1.0, 2, 2, 2, 2, 4, 384)} & 0.95 & 77.86 \\
     
        \bottomrule
        \end{tabular}
        
    \end{subtable}
    \hspace{1.7cm}
    \begin{subtable}[t]{0.4\linewidth}
    \caption{Candidates in $\mathcal{S}_{T_l}$. We denotes $^{1,2,3}$ as Auto-Spikformer $\mathcal{S}_{T_l1,2,3}$}
        \captionsetup{justification=centering}
        \begin{tabular}{ccc}
        \toprule
             \tabincell{c}{Candidates\\ (depth (d), MLP ratio $\times$ d, head num $\times$ d,\\ threshold $\times$ d, tau $\times$ d, time-step, embed dim)} & \tabincell{c}{Energy \\($\mu J$)}  & \tabincell{c}{Acc \\ (\%)}\\
        \midrule
 (2, 4.8, 3.2, 6, 6, 1.4, 1.6, 5, 5, 4, 384) & 0.619 & 77.15 \\
(2, 3.8, 3.2, 6, 12, 0.8, 1.2, 5, 5, 4, 432) & 0.647 & 77.59 \\
(2, 3.8, 4.2, 12, 6, 1.4, 1.4, 3, 5, 4, 432) & \textbf{0.737} & \textbf{77.88} \\
 (2, 3.8, 4.2, 12, 12, 0.8, 1.2, 3, 1.5, 4, 432) & \textbf{0.829} & \textbf{78.08} \\
 (3, 3.2, 3.6, 3.2, 6, 12, 6, 1.4, 1.8, 0.8, 3, 5, 5, 4, 480) & \textbf{0.925} & \textbf{78.22} \\
        \midrule

(2, 3.8, 3.2, 6, 12, 0.8, 1.2, 3, 1.5, 4, 432) $^1$ & \textbf{0.826} & \textbf{78.01} \\ 
(2, 3.8, 4.2, 12, 12, 0.6, 0.8, 3, 1.5, 4, 432) & \textbf{0.829} & \textbf{78.08} \\ 
(2, 4.2, 4.2, 12, 12, 0.8, 1.2, 3, 1.5, 4, 480) $^2$ & \textbf{0.889} & \textbf{78.05} \\ 
(3, 3.2, 3.6, 3.2, 6, 12, 6, 1.4, 1.8, 0.8, 3, 5, 5, 4, 480) $^3$ & \textbf{0.925} & \textbf{78.22} \\ 
(3, 3.2, 3.6, 3.0, 6, 12, 12, 1.4, 2.0, 2.0, 3, 5, 3, 4, 480) & \textbf{0.934} & \textbf{78.17} \\ 
        \midrule
        
 \tabincell{c}{(4, 4, 4, 4, 4, 12, 12, 12, 12,\\1.0, 1.0, 1.0, 1.0, 2, 2, 2, 2, 4, 384)} & 0.95 & 77.86 \\
     
        \bottomrule
        \end{tabular}
        
    \end{subtable}

\end{table}

\subsection{Effectiveness of ESNN}

We train Auto-Spikformer within the SNN search space ($\mathcal{S}_{S}$), where only the SNN parameter sets are modified while maintaining the original Spikformer structure depicted in Table \ref{tab-searchspace} (b). We select 300 candidates through the proposed ESNN and the $\mathcal{F}_{AEB}$. We then plot energy and accuracy for each candidate and draw a Pareto frontier. Notably, by solely modifying the SNN inner parameter sets, a superior trade-off between energy consumption and accuracy can be achieved.

There is a moderate Kendall’s tau rank correlation of 0.4 between the accuracy and the energy consumption. Some candidates exhibit lower energy consumption but higher accuracy, indicating that they are more optimal than others. The energy consumption within $\mathcal{S}_{S}$ is mainly determined by the firing rate, as the architecture is fixed.
We select the candidates located on the Pareto frontier, as well as a subset of candidates with the top 20\% accuracy, and present them in Table \ref{tab-snnspace_cand}. 

We observe that our fitness function and search algorithm favor a time-step of 4, which is the maximum value in $\mathcal{S}_{S}$. Furthermore, we aim to understand why different levels of energy consumption can result in similar accuracy. We notice that the network weights of these candidates are identical. Among them, the minimum energy consumption recorded is 0.79, while the maximum energy consumption is 0.99, resulting in a 25\% difference. Remarkably, despite this significant divergence in energy consumption, the corresponding accuracies achieved are nearly equivalent. 

As shown in Table \ref{tab-snnspace_cand}, for a similar threshold value, the firing rate decreases as the decay parameter increases. The evolutionary search tends to adjust the tau parameter rather than the threshold to control the firing rate. The decay parameter in SNN has a profound effect on the firing rate by facilitating a memory effect for the previous membrane potential. Additionally, the decay and threshold parameters also affect the distribution of feature maps across the layers. Thus, by adjusting the tau and threshold values of each neuron, we can alter the firing rate and accuracy substantially. This shows that the proposed ESNN is a promising approach. By designing an appropriate search space and selecting a suitable fitness function, we are able to effectively decrease the overall firing rate while preserving the network's performance.

\subsection{Effectiveness of $\mathcal{F}_{AEB}$}
\label{sec-joint}
To demonstrate the superiority of the $\mathcal{F}_{AEB}$, we conduct extensive experiments and illustrate the trade-off between energy and accuracy. We apply evolutionary search with the $\mathcal{F}_{AEB}$ as the fitness function to generate 1000 samples in both $\mathcal{S}_{T_s}$ and $\mathcal{S}_{T_l}$. Then we select the top 100 candidates based on their scores. For comparison, we also randomly sample 100 candidates from the search space. Additionally. we also include the Spikformer architecture in the energy-accuracy plot. 

As shown in Figure \ref{fig-stspace-ae}, the Pareto front of the $\mathcal{F}_{AEB}$ dominates the random sample approach. The Kendall’s tau rank correlation coefficients of evolutionary search and random sample are 0.63 and 0.08 in $\mathcal{S}_{T_s}$ and 0.60 and 0.24 in $\mathcal{S}_{T_l}$, respectively. The candidates on the Pareto front are listed in Table \ref{tab-stspace_cand} (a) and the remaining ones are provided in the supplementary materials. We observe that there are numerous candidates that achieve a favorable balance between accuracy and energy consumption. In $\mathcal{S}_{T_s}$, some candidates on the frontier even surpass the original 4-384 Spikformer architecture in accuracy with only 2 blocks and 348 channels, which means half of the energy consumption. $\mathcal{S}_{T_l}$ is used to further explore higher accuracy architecture as shown in Table \ref{tab-stspace_cand} (b). The highest accuracy is 78.22 with a lower energy consumption of 0.925 $\mu J$. Furthermore, several candidates exhibited 10\% to 25\% less energy while achieving higher accuracy compared to the 4-384 Spikformer architecture.

\begin{table}[t!]
\small
\caption{Performance comparison of Auto-Spikformer with existing methods on CIFAR10/100. Auto-Spikformer architectures $\mathcal{S}_{T_s1,2}$ and $\mathcal{S}_{T_l1,2,3}$ are selected from the candidate architectures listed in Table \ref{tab-stspace_cand}. Auto-Spikformer is the first transformer model designed through automated methods, demonstrating enhanced performance in both tasks. The symbol "*" denotes results obtained from self-implemented experiments by \cite{deng2022temporal}. }
\label{tab-cifar}
\begin{center}
\setlength{\tabcolsep}{3pt}
\resizebox{1\textwidth}{!}{
\begin{tabular}{ccccccccc}
\toprule
  \multicolumn{1}{c}{\bf Methods} &\multicolumn{1}{c}{\bf Architecture} &\multicolumn{1}{c}{\bf \tabincell{c}{Param (M)\\/ Energy ($\mu J$)}}
&\bf\tabincell{c}{Time\\Step} &\bf\tabincell{c}{CIFAR10\\Acc} &\bf\tabincell{c}{CIFAR100\\Acc}
& \bf\tabincell{c}{Model \\Type} & \bf\tabincell{c}{Design \\Type}
\\
\midrule
    Hybrid training\citep{rathi2020enabling} &VGG-11 &9.27 &125 &92.22 &67.87 & CNN & Manual\\
    Diet-SNN\citep{rathi2020diet} &ResNet-20 &0.27 &10\textbf{/}5  & 92.54& 64.07 & CNN & Manual\\
    STBP\citep{wu2018spatio} &CIFARNet &17.54&12 & 89.83&- & CNN & Manual\\
    STBP NeuNorm\citep{wu2019direct} &CIFARNet &17.54 &12 &90.53& - & CNN & Manual\\
    TSSL-BP\citep{zhang2020temporal} &CIFARNet &17.54 &5 &91.41& - & CNN & Manual\\
    \multicolumn{1}{c}{\multirow{1}{*}{{STBP-tdBN\citep{zheng2021going}}}} &\multicolumn{1}{c}{\multirow{1}{*}{ResNet-19}} &12.63 & 4 & 92.92 & 70.86 & CNN & Manual\\
    TET\citep{deng2022temporal}  &\multicolumn{1}{c}{\multirow{1}{*}{ResNet-19}} &12.63 & 4 & {94.44}& {74.47} & CNN & Manual\\
    AutoSNN\cite{na2022autosnn} & AutoSNN (C=128) & 21 & 8 & 93.15 & 69.16 & CNN & Auto\\
    SNASNet\cite{kim2022neural} & SNASNet-Bw & - & 8 & 94.12 & 73.04 & CNN & Auto\\
    SpikeDHS$^D$ \cite{che2022differentiable} & SpikeDHS-CLA (n3s1) & 14 & 6 & 95.36 & 76.25 & CNN & Auto\\
\midrule
\multicolumn{1}{c}{\multirow{2}{*}{\textbf{ANN}}} 
     &ResNet-19* &12.63 &1 &{94.97} & {75.35} & CNN & Manual\\
     & {Transformer-4-384} & {9.32} / 3.97 &1 &{\textbf{96.73}} & {\textbf{81.02}} & Transformer & Manual\\
\midrule
    \multicolumn{1}{c}{\multirow{3}{*}{\textbf{Spikformer}}}  
    &\multicolumn{1}{c}{\multirow{1}{*}{Spikformer-4-256}}&4.15 / 0.553 & 4 & 93.94 & {75.96} & Transformer & Manual\\
    &\multicolumn{1}{c}{\multirow{1}{*}{Spikformer-2-384}}&5.76 / 0.582 & 4 & {94.80} & {76.95} & Transformer & Manual\\
    &\multicolumn{1}{c}{\multirow{1}{*}{Spikformer-4-384}}&9.32 / 0.952 & 4 & {95.19} & {77.86} & Transformer & Manual\\
\midrule
    \multicolumn{1}{c}{\multirow{5}{*}{\textbf{Auto-Spikformer}}}  
    &\multicolumn{1}{c}{\multirow{1}{*}{Auto-Spikformer $\mathcal{S}_{T_s1}$}}& 4.69 / \textbf{0.505} & 4 & \textbf{95.29} & \textbf{77.71} & Transformer & Auto\\
    &\multicolumn{1}{c}{\multirow{1}{*}{Auto-Spikformer $\mathcal{S}_{T_s2}$}}& 4.64 / \textbf{0.509} & 4 & \textbf{95.23} & \textbf{77.91} & Transformer & Auto\\
    &\multicolumn{1}{c}{\multirow{1}{*}{Auto-Spikformer $\mathcal{S}_{T_l1}$}}& \textbf{7.09} / \textbf{0.826}  & 4 & \textbf{96.19} & \textbf{78.01} & Transformer & Auto\\
    &\multicolumn{1}{c}{\multirow{1}{*}{Auto-Spikformer $\mathcal{S}_{T_l2}$}}& \textbf{9.20} / \textbf{0.889}  & 4 & \textbf{96.38} & \textbf{77.05} & Transformer & Auto\\
    &\multicolumn{1}{c}{\multirow{1}{*}{Auto-Spikformer $\mathcal{S}_{T_l3}$}}& \textbf{8.46} / \textbf{0.925} & 4 & \textbf{96.39} & \textbf{78.22} & Transformer & Auto\\

\bottomrule
\end{tabular}}
\end{center}

\vspace{-6mm}
\end{table}

\subsection{Result on CIFAR}
We select the Auto-Spikformer architecture searched in $\mathcal{S}_{T_s}$ and $\mathcal{S}_{T_l}$ in Section \ref{sec-joint} and compare with original Spikformer and other methods. The performances are reported in Table \ref{tab-cifar}. Auto-Spikformer is the first transformer model designed through automated methods.
Auto-Spikformer $\mathcal{S}{T_s2}$ and $\mathcal{S}{T_l1,2,3}$ outperform the state-of-the-art method including CNN or Transformer models that are manually or automatically designed in both accuracy and energy consumption. Compare to the original Spikformer, the Auto-Spikformer obtains a significant improvement of accuracy with even less energy consumption. The ANN-Transformer model is only 0.34\% and 2.8 \% than $\mathcal{S}{T_l1,2,3}$, respectively, which demonstrates that the Auto-Spikformer method is comparable to the ANN version.

\section{Conclusion}


In this work, we are the first to propose a one-shot architecture search method for spiking-based vision transformers, called Auto-Spikformer. Auto-Spikformer optimizes both energy consumption and accuracy by incorporating key parameters of SNN and transformer into the search space. We propose two novel methods: Evolutionary SNN neurons (ESNN), which optimizes the SNN parameters, and Accuracy and Energy Balanced Fitness Function $\mathcal{F}_{AEB}$, which balances the energy consumption and accuracy objectives. Extensive experiments show that the proposed algorithm can improve the performance of Spikformer and discover many promising architectures. As a future work, we plan to conduct experiments on larger benchmark datasets or neuromorphic datasets.

\bibliographystyle{plainnat}
\bibliography{neurips_bib.bib}

\end{document}